\documentclass[runningheads]{llncs}
\usepackage{graphicx}
\usepackage{hyperref}
\usepackage{multirow}
\usepackage{algorithm}
\usepackage{algorithmic}
\usepackage{amsmath,amsfonts,amssymb}
\usepackage[misc,geometry]{ifsym}
\usepackage[T5]{fontenc}
\usepackage[utf8]{inputenc}
\DeclareUnicodeCharacter{2061}{}
\begin{document}
\title{Analyzing Vietnamese Legal Questions using Deep Neural Networks with Biaffine Classifiers}
\titlerunning{Analyzing Vietnamese Legal Questions using Deep Neural Networks}
%
\author{Nguyen Anh Tu\inst{1,2} \and Hoang Thi Thu Uyen\inst{1} \and \\ Tu Minh Phuong\inst{1} \and Ngo Xuan Bach\inst{1}\textsuperscript{(\Letter)}}
\authorrunning{N.A. Tu et al.}
%
\institute{Department of Computer Science, Posts and Telecommunications Institute of Technology, Hanoi, Vietnam
\and
FPT Technology Research Institute, FPT University, Hanoi, Vietnam
\email{\{anhtunguyen446,thuuyenptit\}@gmail.com; \{phuongtm,bachnx\}@ptit.edu.vn}}
\maketitle              
\begin{abstract}
In this paper, we propose using deep neural networks to extract important information from Vietnamese legal questions, a fundamental task towards building a question answering system in the legal domain. Given a legal question in natural language, the goal is to extract all the segments that contain the needed information to answer the question. We introduce a deep model that solves the task in three stages. First, our model leverages recent advanced autoencoding language models to produce contextual word embeddings, which are then combined with character-level and POS-tag information to form word representations. Next, bidirectional long short-term memory networks are employed to capture the relations among words and generate sentence-level representations. At the third stage, borrowing ideas from graph-based dependency parsing methods which provide a global view on the input sentence, we use biaffine classifiers to estimate the probability of each pair of start-end words to be an important segment. Experimental results on a public Vietnamese legal dataset show that our model outperforms the previous work by a large margin, achieving 94.79\% in the F$_1$ score. The results also prove the effectiveness of using contextual features extracted from pre-trained language models combined with other types of features such as character-level and POS-tag features when training on a limited dataset.
\end{abstract}

\section{Introduction}
Question answering (QA) \cite{Duong:15,Kien:20,Nguyen:09,Tran:09,Tran:12,Tran:14}, a sub-field of natural language processing (NLP) and information retrieval (IR), aims to build computer systems that can automatically answer questions in natural languages. There are two main approaches for building QA systems: IR-based and knowledge-based. The former approach finds and extracts answers from a collection of unstructured text, while the latter utilizes structured text or knowledge bases. Although two approaches exploit different types of resources, they share the first step, question analysis, which extracts the needed information to answer the input question. Such information is then used to form a query in various ways which serves as the input for the next step. Question analysis is therefore a crucial task for both IR-based and knowledge-based question answering.  

\begin{figure}[t]
	\begin{center}
		\includegraphics[width = 11.3cm]{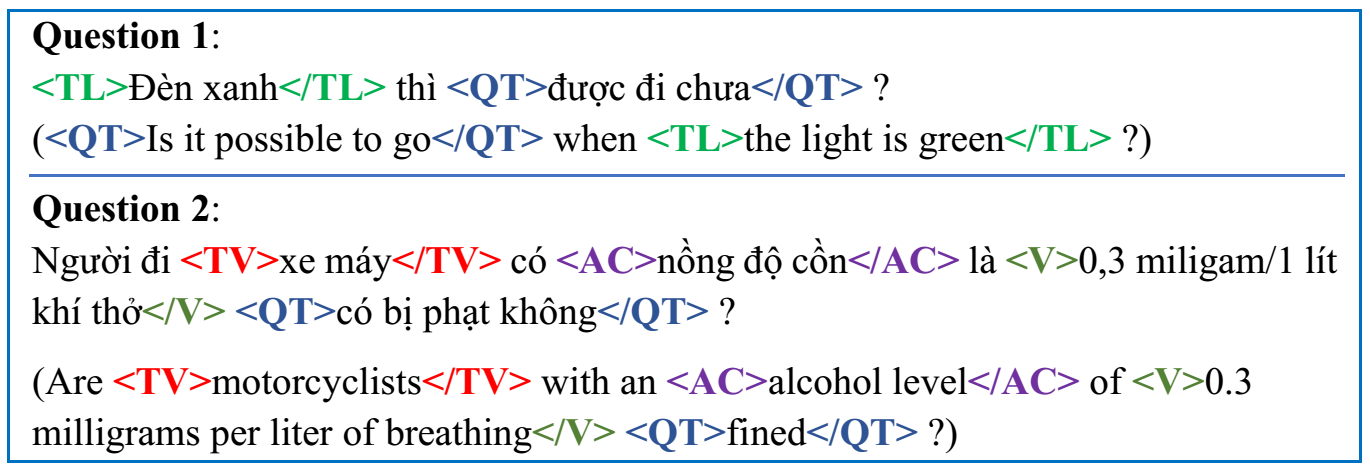}
	\end{center}	
		\caption{Examples of two legal questions and their important information.}
	\label{fig:examples}
\end{figure}

In this paper, we target at the question analysis task in the legal domain, which is undoubtedly important but has received less attention. Legal and regulatory documents are ubiquitous and have a great impact on our life. Figure \ref{fig:examples} shows two examples of Vietnamese legal questions annotated with key information. The goal is to correctly extract two types of information in the first question (Traffic Light-TL and Question Type-QT), and four types of information in the second question (Type of Vehicle-TV, Alcohol Concentration-AC, Value-V, and Question Type-QT). We call the segments that contain important information important segments.  

Traditional methods often frame the task as a sequence labeling problem and exploit probabilistic graphical models like conditional random fields (CRFs) \cite{Lafferty:01} to solve it. The main advantage of those methods is that they can build an accurate model using a relatively small annotated corpus with a handcrafted feature set. Recently, however, deep neural networks have made tremendous breakthroughs in various NLP tasks and applications, including sentence classification \cite{Kim:14,Song:18}, sequence labeling \cite{Bach:19,He:20,Yadav:18}, syntactic and dependency parsing \cite{Mrini:20,Yang:20}, natural language inference \cite{Devlin:19,He:21}, machine translation \cite{Sutskever:14,Yang:20MT}, as well as question answering \cite{He:21,Kien:20}. Furthermore, the well-known limitation of deep models, i.e. data hungry, could be mitigated by using advanced pre-trained models and fine-tuning \cite{Devlin:19,He:21}. All of these make deep neural networks the preferred choice for NLP tasks in general and QA in particular.   

Here, we propose to use deep neural networks for Vietnamese question analysis in the legal domain. Our method combines several recent advanced techniques in the NLP and deep learning research communities: pre-trained language models \cite{Devlin:19} for contextual word embeddings, convolutional neural networks (CNNs) \cite{LeCun:98} for extracting character-level features, and bidirectional long-short term memory networks (BiLSTM) \cite{Graves:05} for sentence-level representations. Furthermore, instead of formulating it as a sequence labeling problem, we employ biaffine classifiers to estimate directly the possibility that a pair of words becomes an important segment. The main advantage of biaffine classifiers is that they provide a global view on the input sentence, which has been shown to be effective in dependency paring \cite{Dozat:17,Li:19}, and named entity and relation extraction \cite{Nguyen:19,Yu:20}. Experimental results on a Vietnamese corpus consisting of 1678 legal questions show that our model outperforms a SOTA method by a large margin, showing the F$_1$ score of 94.79\%. The effectiveness of these components of the model is also validated by an ablation study.

The remainder is organized as follows. Section \ref{sec:related} reviews related work. Section \ref{sec:method} presents our model for extracting important information from Vietnamese legal questions. The model architecture is presented first, and its key components are then described in more detail. Experimental results and error analysis are introduced in Section \ref{sec:experiments}. Finally, Section \ref{sec:conclusion} concludes the paper and shows some future work. 

\section{Related Work}
\label{sec:related}

Several studies have been performed on Vietnamese QA in various domains, including travel, education, as well as legal. Tran et al. \cite{Tran:09} introduce a Vietnamese QA system, which can answer simple questions in the travel domain by mining information from the Web. Bach et al. \cite{Bach:20} focus on the task of analyzing Vietnamese question in education. Using deep neural networks (BiLSTM-CRFs) with a rich set of features, their model can accurately extract 14 types of vital information from education utterances. In the legal domain, Duong and Ho  \cite{Duong:15} develop a QA system to answer Vietnamese questions about provisions,  procedures, processes, etc. in enterprise laws. Kien et al. \cite{Kien:20} introduce a retrieval-based method for answering Vietnamese legal questions by learning text representation. Their model leverages CNNs and attention mechanisms to extract and align important information between a question and a legal article. Other works on Vietnamese question answering include Nguyen et al. \cite{Nguyen:09}, Tran et al. \cite{Tran:12}, and Le-Hong and Bui \cite{Le:18}.  

Perhaps the most closely work to ours is the one of Bach et al. \cite{Bach:17}, which also focuses on analyzing Vietnamese legal questions. Our method, however, is distinguished from theirs in two aspects. First, we formulate the task as a multi-class classification problem instead of sequence labeling. Second, we utilize deep neural networks instead of using traditional models like CRFs. 

\section{Method}
\label{sec:method}

\subsection{Model Overview}
\begin{figure}[t]
	\begin{center}
		\includegraphics[width = 12cm]{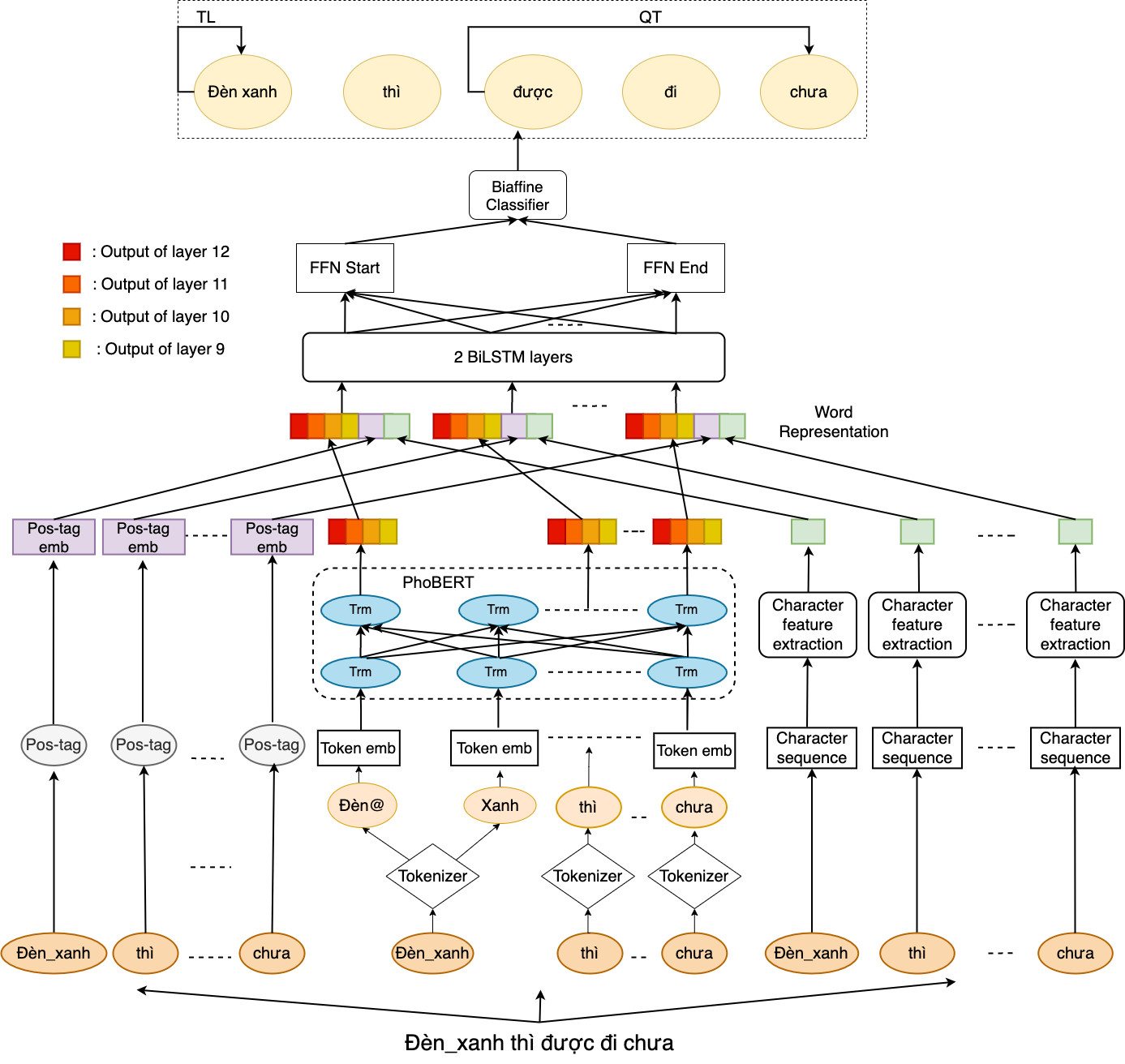}
	\end{center}	
	\caption{Model architecture.}
	\label{fig:model}
\end{figure}

Our goal is to extract all important segments from an input legal question, where each segment is a triple of start/end positions and a label for the information type. Figure \ref{fig:model} illustrates our proposed architecture. First, we create word representations by concatenating different types of features: contextual word embeddings, character-level features, and part-of-speech (POS) tag embeddings. The outputs of the word representation layer are then fed into two stacked BiLSTMs to obtain the sentence-level representations. After that, we use two feed forward neural networks (FFN) to generate different representations for the start/end of segments. Finally, we employ a biaffine classifier to create a  $n \times n \times c$ scoring tensor $R$, where $n$ denotes the length of the input question, and $c$ is the number of labels (including Null for non important information). Tensor $R$ provides scores for all possible segments that could contain key information. 

In the up-coming sections, we show the model components in detail. For notation, we denote vectors, matrices, and scalars   with bold lower-case (e.g., $\textbf{x}_t$, $\textbf{h}_t$, $\textbf{b}$), bold upper-case (e.g., $\textbf{H}$, $\textbf{W}_i$, $\textbf{V}_i$), and italic lower-case (e.g., $n$, $c$), respectively.

\subsection{Word Representations}
Because words are basic elements to form written languages, a good word representation method is the first and crucial step to build successful NLP systems. In this work, we create rich information word representations by integrating multiple information sources, including contextual word embeddings, character-level features, and POS-tag embeddings.

\textbf{Contextual word embeddings}.
Traditional NLP approaches usually represent words by one-hot vectors with one value 1 and the rest 0. These high-dimensional sparse vectors are memory consuming and cannot capture the word semantics. Distributed word representation methods, which use low-dimensional continuous vectors, have been introduced to handle these issues.  Word2vec \cite{Mikolov:13}, Fasttext \cite{Bojanowski:17}, and Glove \cite{Pennington:14} are successful examples of such methods that represent similar words with similar vectors. Although these methods have made many breakthroughs in NLP research, they represent words by fix vectors which are context independent. Static word embedding methods like word2vec are therefore limited in representing polysemous words.    

Recently, contextual word embedding methods have been shown to be the key component of many SOTA NLP systems \cite{Devlin:19,Liu:19}. The main advantage of these methods it that they can learn different representations for polysemous words by considering the sequence of all words in sentences/documents. Perhaps BERT proposed by Devlin et al. \cite{Devlin:19} is the most famous and popular contextual word embedding method. The key technical innovation of BERT is applying the bidirectional training of Transformer \cite{Vaswani:17} to language modeling with two strategies: masked-language modeling and next sentence prediction. 

In this work, we use PhoBERT \cite{Nguyen:20}, a monolingual variant of RoBERTa \cite{Liu:19} pre-trained on a 20GB word-level Vietnamese dataset. Like BERT and RoBERTa, PhoBERT segments the input sentence into sub-words, which brings the balance between character- and word-level hybrid representations and enables the encoding of rare words with appropriate sub-words. We represent each sub-word by concatenating embeddings of the last four encoding layers (9 to 12) of PhoBERT-base, and the contextual embedding of a word is the embedding of its first sub-word. 

\begin{figure}[t]
	\begin{center}
		\includegraphics[width = 8.5cm]{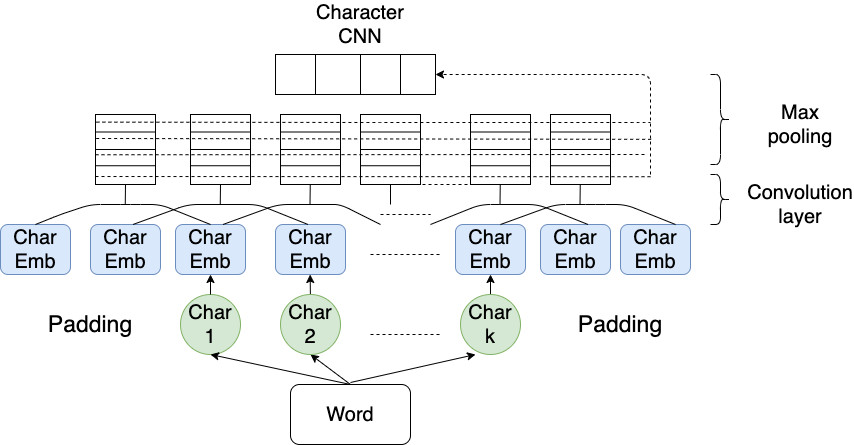}
	\end{center}	
		\caption{CNN-based character-level features.}
	\label{fig:CharCNN}
\end{figure}

\textbf{Character-level features}.
Beside contextual word embeddings, we also utilize morphological information from characters. Additional character embeddings are derived from character-level convolutional (charCNN) networks. As shown in Figure \ref{fig:CharCNN}, charCNN consists of 1D operations: convolution and max pooling. Feature maps are then concatenated to produce a character-level representation for the word.

\textbf{POS-tag embeddings}.
We suppose POS tags are another useful source of information. Therefore, we also use POS tag embeddings to represent words. These embedding vectors are initialized randomly in range $(-\sqrt{3/dim},\sqrt{3/dim})$, where $dim$ denotes their dimension. We use VnCoreNLP \cite{Vu:18} for word segmentation and POS tagging for input questions. 

Finally, all feature vectors are concatenated into a single embedding for representing a word. Word vectors are then fed into BiLSTM networks to create sentence-level representations. 

\subsection{BiLSTM}
Long short-term memory (LSTM) networks \cite{Hochreiter:97} are designed for sequence data modeling problem. Let $\textbf{X} = (\textbf{x}_1,\textbf{x}_2,\ldots,\textbf{x}_n)$ denote the input question where $\textbf{x}_i$ is the embedding of the $i^{th}$ word. At each position $t$, the LSTM computes an intermediate representation using a hidden state $\textbf{h}$:
$$\textbf{h}_t=f(\textbf{h}_{t-1},\textbf{x}_t)$$
where $f$ includes an input gate, a forget gate, an output gate, and a memory cell (denoted by $\textbf{i}_t$, $\textbf{f}_t$, $\textbf{o}_t$, $\textbf{c}_t$, respectively) to update $\textbf{h}_t$:
$$\textbf{i}_t = \sigma(\textbf{W}_i \textbf{x}_t + \textbf{V}_i \textbf{h}_{t-1} + \textbf{b}_i),$$
$$\textbf{f}_t = \sigma(\textbf{W}_f \textbf{x}_t + \textbf{V}_f \textbf{h}_{t-1} + \textbf{b}_f),$$
$$\textbf{o}_t = \sigma(\textbf{W}_o \textbf{x}_t + \textbf{V}_o \textbf{h}_{t-1} + \textbf{b}_o),$$
$$\textbf{c}_t = \textbf{f}_t \odot \textbf{c}_{t-1} + \textbf{i}_t \odot \text{tanh}⁡(\textbf{W}_c \textbf{x}_t + \textbf{V}_c \textbf{h}_{t-1} + \textbf{b}_c),$$
$$\textbf{h}_t = \textbf{o}_t \odot \text{tanh}⁡(\textbf{c}_t),$$
and the output $\textbf{y}_t$ can be produced based on $\textbf{h}_t$:
$$\textbf{y}_t = \sigma(\textbf{W}_y \textbf{h}_t + \textbf{b}_y),$$
where $\odot$ indicates the multiplication operator function, $\sigma$ is the element-wise softmax, and $\textbf{W}_*$, $\textbf{V}_*$, and $\textbf{b}_*$ ($*$ denotes $i$,$f$,$o$,$c$,$y$) are weight matrices and vectors to be learned during the training process. 

Bidirectional long short-term memory (BiLSTM) \cite{Graves:05} combine two LSTMs: one network moves from the left to the right and the other network moves from the right to the left of the sequence. Two BiLSTMs are exploited to learn a higher level of semantics within the sentence. 

\subsection{Biaffine Layer}
We employ two feed forward neural networks (FFNs) to create different representations for start/end positions. The outputs of two FFNs at position $t$ are denoted by $\textbf{g}_t^{start}$  and $\textbf{g}_t^{end}$: 
$$\textbf{g}_t^{start} = \text{FFN}^{start}(\textbf{y}_t)$$
$$\textbf{g}_t^{end} = \text{FFN}^{end}(\textbf{y}_t)$$
For each start-end candidate pair $(i,j)$, $1\leq i\leq j\leq n$, we apply the biaffine classifier:
$$\textbf{r}_{i,j}=\text{Biaffine}(\textbf{g}_i^{start},\textbf{g}_j^{end})= (\textbf{g}_i^{start})^{\top} \textbf{U} \textbf{g}_j^{end} + \textbf{W}(\textbf{g}_i^{start}\oplus \textbf{g}_j^{end}) + \textbf{b},$$
where $\textbf{U}$, $\textbf{W}$, $\textbf{b}$ are a $d\times c\times d$ tensor, a $c\times 2d$ matrix, and a bias vector, respectively, and $d$ is the size of the output layers of both FFN$^{start}$ and FFN$^{end}$.

Vector $\textbf{r}_{i,j}$ is then fed into a softmax layer to produce probability scores $\textbf{s}_{i,j}$:
$$\textbf{s}_{i,j}(k)=\frac{exp⁡(\textbf{r}_{i,j}(k))}{\sum_{k'=1}^{c}{exp⁡(\textbf{r}_{i,j}(k^{'}))}} $$
The label of segment $(i,j)$ can be determined as: $\widehat{l} = \arg \max_{k} \textbf{s}_{i,j}(k).$

The question analysis task now becomes a multi-class classification problem and model parameters are learned to minimize the cross-entropy loss function. 

\section{Experiments}
\label{sec:experiments}

\subsection{Data and Evaluation Method}
In our experiments we used the Vietnamese dataset of legal questions introduced by Bach et al. \cite{Bach:17}. This dataset consists of 1678 legal questions about the traffic law in Vietnam. Questions were annotated with 16 labels reflecting different aspects of the domain listed in Table \ref{tab:dataset}.
\begin{table}[tbp]
	\centering
	\caption{Types of important information and their occurrence numbers}
		\label{tab:dataset}
		\begin{tabular}{c|l|c|c|l|c}
			\hline	
			\hline
				\textbf{Label} &	\textbf{Meaning} &	\textbf{\#} &\textbf{Label} &	\textbf{Meaning} &	\textbf{\#}\\
			\hline
			\hline
				A	& Action of vehicle &	1087 & L &	Location &	426\\
				\hline
				AC &	Alcohol concentration &	44 & QT &	Question type &	1678\\
				\hline
				ANO &	Annotation &	75 & SP &	Speed &	115\\
				\hline
				DL &	Driving license &	119 & TI &	Traffic instructor &	93\\
				\hline
				IF1 &	Add. info. about vehicle &	196 & TL &	Traffic light &	31\\
				\hline
				IF2 &	Add. info. about traffic light &	12 & TP &	Traffic participant &	20\\
				\hline
				IF3 &	Add. info. about traffic participant &	287 & TV &	Type of vehicle &	1245\\
				\hline
				IF4	& Add. info. &	227 & V &	Value &	231\\				
			\hline
			\hline
			\end{tabular}	
\end{table}

We performed cross-validation tests with the same training/test data splits of Bach et al. \cite{Bach:17}. For each fold, we used $10$\% of the training set as a validation set. The performance of extraction models was measured using popular metrics such as precision, recall and the F$_1$ score. 

\subsection{Network Training}
Our models were implemented in PyTorch\footnote{\url{https://pytorch.org/}} using Huggingface's Transformers\footnote{\url{https://huggingface.co/transformers/}}. In all experiments, we set the batch size to $64$. The max character length was set to $15$, and the max sequence length was tuned in [$50, 60, 80, 100$] for all models, and the best value was $60$. We set the dimensions of character-level features and POS-tag embeddings to $256$ and $100$, respectively. We used dimension of $300$ for FFNs, and kernel sizes of 3 and 4 for charCNN. To mitigate overfitting, we used a dropout rate of $0.3$ for each hidden layer. Our models were trained using the AdamW optimizer \cite{Loshchilov:19}. We set the epsilon and weight decay to default values in PyTorch, i.e. 1$e$-8. The learning rate was tuned in [3$e$-5, 4$e$-5, 5$e$-5] and the best learning rate value was 5$e$-5. For each model, we trained for $30$ epochs and selected the version that obtained the highest F$_1$ score on the validation set to apply to the test set. 

\subsection{Experimental Results}
\subsubsection{Our Model vs. Baseline}
We first conducted experiments to compare our model with the previous work of Bach et al. \cite{Bach:17} (our baseline). Table \ref{tab:results} shows experimental results on each type of information and the overall scores of two models. Our model outperformed the baseline for 15 out of 16 types of information. Types with the biggest improvements in the F$_1$ score include TL (Traffic light: 19.79\%), ANO (Annotation: 9.61\%), TP (Traffic participant: 7.48\%), A (Action: 4.47\%), and SP (Speed: 4.33\%). Overall, our model achieved a micro F$_1$ score of 94.79\%, which improved 1.85\% (26.20\% error rate reduction) in comparison with the baseline. Experimental results demonstrated the effectiveness of deep neural networks compared to traditional methods like CRFs.        
\begin{table}[tbp]
	\centering
	\caption{Experimental results of our model compared with the baseline (the improvements are indicated in bold)}
		\label{tab:results}
		\begin{tabular}{c|c|c|c|c|c|c}
			\hline	
			\hline	
						& \multicolumn{3}{|c|}{\textbf{Baseline} (Bach et al. \cite{Bach:17})}	& \multicolumn{3}{|c}{\textbf{Our Model}}\\
			\hline
			\hline
			\textbf{Type} &	\textbf{Prec.(\%)} &	\textbf{Rec.(\%)} &	\textbf{F$_1$(\%)} &	\textbf{Prec.(\%)} &	\textbf{Rec.(\%)} &	\textbf{F$_1$(\%)}\\	
			\hline
			\hline
			A	& 88.20 &	89.14 &	88.66 &	92.91 &	93.35 &	93.13 (\textbf{4.47}$\uparrow$)\\
			\hline
			AC &	95.78 &	95.78 &	95.78 &	96.12 &	97.28 &	96.70 (\textbf{0.92}$\uparrow$)\\
			\hline
			ANO	& 85.58 &	60.57 &	68.82 &	73.17 &	84.51 &	78.43 (\textbf{9.61}$\uparrow$)\\
			\hline
			DL &	97.97 &	99.20 &	98.54 &	100.00 &	98.26 &	99.12 (\textbf{0.58}$\uparrow$)\\
			\hline
			IF1 &	94.64 &	87.35 &	90.67 &	90.34 &	91.48 &	90.91 (\textbf{0.24}$\uparrow$)\\
			\hline
			IF2 &	100.00 &	73.33 &	82.67 &	86.72 &	70.44 &	77.74 (4.93$\downarrow$)\\
			\hline
			IF3 &	88.08 &	75.06 &	80.91 &	85.77 &	83.45 &	84.59 (\textbf{3.68}$\uparrow$)\\
			\hline
			IF4 &	85.77 &	74.34 &	79.51 &	80.14 &	82.97 &	81.53 (\textbf{2.02}$\uparrow$)\\
			\hline
			L	& 92.23 &	92.71 &	92.44 &	93.18 &	96.31	& 94.72 (\textbf{2.28}$\uparrow$)\\
			\hline
			QT &	96.03 &	94.84 &	95.42 &	95.89 &	96.07 &	95.98 (\textbf{0.56}$\uparrow$)\\
			\hline
			SP &	95.61 &	91.44 &	93.23 &	100.00 &	95.24 &	97.56 (\textbf{4.33}$\uparrow$)\\
			\hline
			TI	& 99.05 &	95.53 &	97.24 &	100.00 &	100.00 &	100.00 (\textbf{2.76}$\uparrow$)\\
			\hline
			TL	& 86.95 &	60.35 &	68.59 & 87.33 &	89.46 &	88.38 (\textbf{19.79}$\uparrow$)\\
			\hline
			TP	& 80.00 &	60.67 &	67.21 &	76.35 &	73.11 &	74.69 (\textbf{7.48}$\uparrow$)\\
			\hline
			TV	& 97.32 &	98.78 &	98.04 &	99.57 &	97.99 &	98.77 (\textbf{0.73}$\uparrow$)\\
			\hline
			V	& 98.18 &	99.26 &	98.71 &	100.00 &	98.23 &	99.11 (\textbf{0.40}$\uparrow$)\\
			\hline
			\hline
			Overall &	93.84 &	92.05 &	92.94 &	94.30 &	95.28 &	94.79 (\textbf{1.85}$\uparrow$)\\
			\hline
			\hline
			\end{tabular}	
\end{table}

\subsubsection{Ablation Study}
Next we evaluated the contribution of individual components of our model by performing an ablation study as follows: 
\begin{itemize}
	\item \textbf{Biaffine classifier}: We replaced the biaffine classifier with a CRF/softmax layer and reformulated the task as a sequence labeling problem
	\item \textbf{Character-level features}, \textbf{POS-tag embeddings}, \textbf{BiLSTMs}: We removed each component in turn.
	\item \textbf{PhoBERT Embeddings}: We replaced contextual word embeddings with static word embeddings. In our experiments, we used Fasttext \cite{Bojanowski:17}, a variant of Word2Vec \cite{Mikolov:13} which deals with unknown words and sparsity in languages by using sub-word models.	
\end{itemize}

Experimental results in Table \ref{tab:ablation} proved that all the components contributed to the success of our model. Our first observation is that the performance of the system degraded when we modified or removed a component. While the full model got 94.79\%, the F$_1$ score reduced to 94.62\% and 93.59\% when we replaced the biaffine classifier with CRF/softmax function and reformulated the task as a sequence labeling problem. The score was only 94.46\%, 94.33\%, and 93.77\%, when we removed POS-tag features, character-level features, and the BiLSTM layers, respectively. The results showed that the BiLSTM layers have a high impact on the performance of our model. The second observation is that replacing contextual word embeddings by static word embeddings leads to the biggest decrease of 1.98\%. This indicated that contextual word embeddings from pre-trained language models like PhoBERT played a critical role in our model.      
\begin{table}[tbp]
	\centering
	\caption{Ablation study (the decrease in the F$_1$ score of the modified models is indicated in bold)}
		\label{tab:ablation}
		\begin{tabular}{l|l|c|c|c}
			\hline	
			\hline
				\textbf{Component} &	\textbf{Modification} &	\textbf{Prec.(\%)}	& \textbf{Rec.(\%)} &	\textbf{F$_1$(\%)}\\
			\hline
			\hline
			Full &	None &	94.30 &	95.28 &	94.79\\ 	
			\hline
			\hline
			\multirow{2}{*}{Biaffine classifier}	& Seq. labeling, CRF &	93.92 & 95.33 & 94.62 (\textbf{0.17}$\downarrow$)\\
			\cline{2-5}
			& Seq. labeling, Softmax &	94.02 & 93.17 & 93.59 (\textbf{1.20}$\downarrow$)\\
			\hline
			POS-tag embeddings &	Removal &	94.54 &	94.39 &	94.46 (\textbf{0.33}$\downarrow$)\\
			\hline
			Character features &	Removal &	93.51 &	95.16	& 94.33 (\textbf{0.46}$\downarrow$)\\
			\hline
			BiLSTM	& Removal &	92.95 &	94.60 &	93.77 (\textbf{1.02}$\downarrow$)\\
			\hline
			PhoBERT	& Fasttext &	92.00 &	93.63 &	92.81 (\textbf{1.98}$\downarrow$)\\
			\hline
			\hline
			\end{tabular}	
\end{table}

\subsection{Error Analysis}
This section discusses the cases in which our model failed to extract important information. By analyzing the model's predictions on the test sets, we found that most errors belong to one of two following types:
\begin{itemize}
	\item \textbf{Type I: Incorrect segments}. Our model identified segments that are shorter or longer than the gold segments. 
	\item \textbf{Type II: Incorrect information types}. Our model recognized segments (start and end positions) correctly but assigned wrong information types (labels) to the segments.  
\end{itemize}

\begin{figure}[tbp]
	\begin{center}
		\includegraphics[width = 10.5cm]{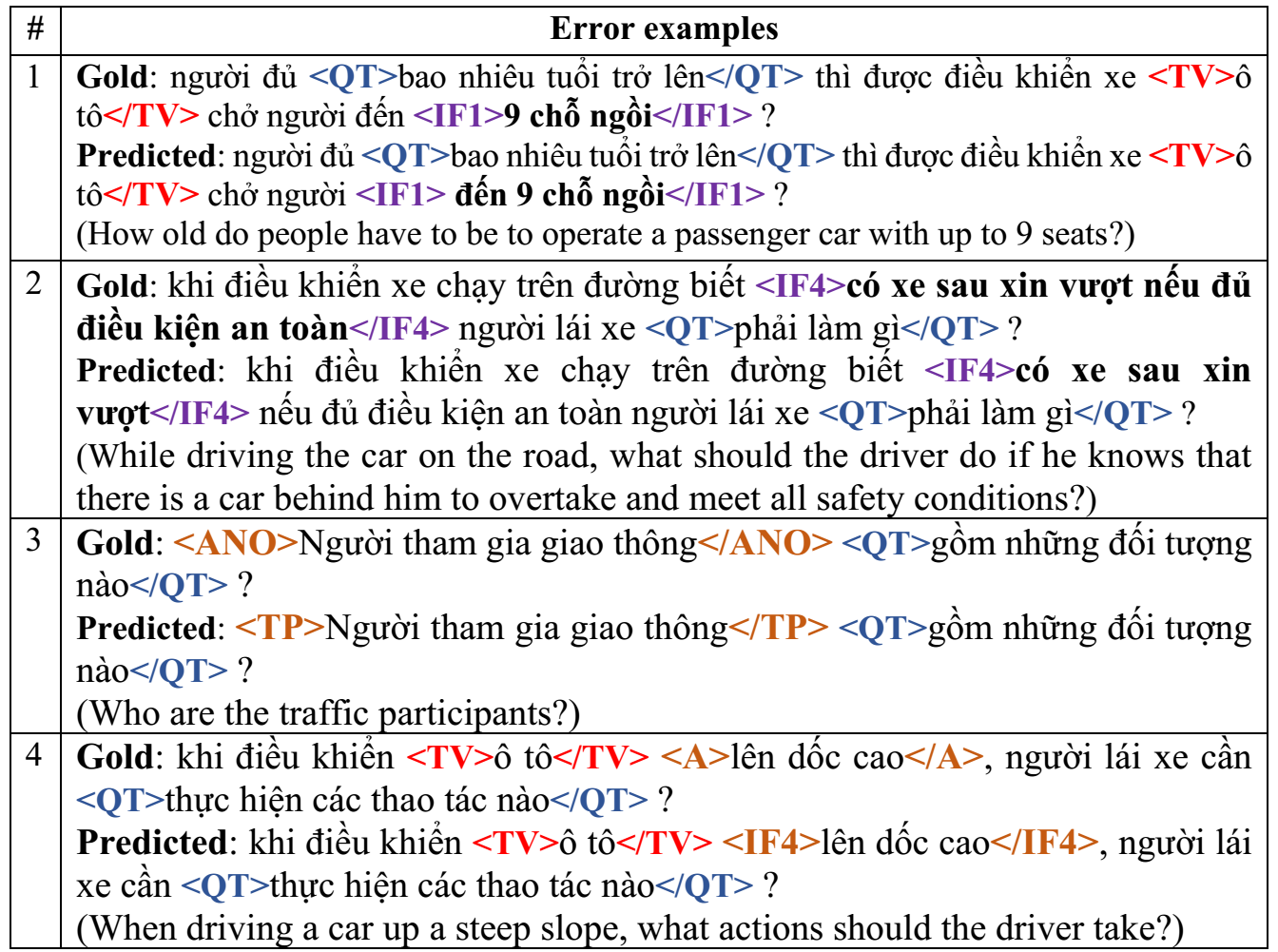}
	\end{center}	
	\caption{Examples of error cases.}
	\label{fig:errors}
\end{figure}

Figure \ref{fig:errors} shows four examples of error cases. Among them, the first two examples belong to Type I and the others belong to Type II. While our model identified a longer segment in the first case, it detected a shorter segment in the second case. In the third and fourth examples, our model made a mistake on labels (TP instead of ANO, and IF4 instead of A).  

\section{Conclusion}
\label{sec:conclusion}
We have introduced a deep neural network model for analyzing Vietnamese legal questions, a key step towards building an automatically question answering system in the legal domain. By utilizing recent advanced techniques in the NLP and deep learning research communities, our model can correctly extract 16 types of important information. For future work, we plan to develop a question answering system for Vietnamese legal questions. Studying deep neural network models for other NLP tasks in Vietnamese language is another direction for future work.  

\section*{Acknowledgements}		
We would like to thank FPT Technology Research Institute, FPT University for financial support which made this work possible.
%
%

\end{document}